\documentclass[final]{cvpr}

\usepackage{times}
\usepackage{epsfig}
\usepackage{graphicx}
\usepackage{amsmath}
\usepackage{amssymb}
\usepackage{subfigure}
\usepackage{overpic}

\usepackage{bm}
\usepackage{algorithm}

\usepackage{algorithmicx}
\usepackage{algpseudocode}

\usepackage{enumitem}
\setenumerate[1]{itemsep=0pt,partopsep=0pt,parsep=\parskip,topsep=5pt}
\setitemize[1]{itemsep=0pt,partopsep=0pt,parsep=\parskip,topsep=5pt}
\setdescription{itemsep=0pt,partopsep=0pt,parsep=\parskip,topsep=5pt}

\usepackage[pagebackref=true,breaklinks=true,colorlinks,bookmarks=false]{hyperref}

\def\cvprPaperID{159} % *** Enter the CVPR Paper ID here
\def\confYear{CVPR 2020}

\graphicspath{{figures/}}

\newcommand{\Ignore}[1]{}

\newcommand{\NullFigure}[3]{}

\title{An Interactive Image-based Modeling System}

\author{papers 0578}
\author{Zhi He ~~~~~ Rui Wang ~~~~~ Wei Hua ~~~~~ Yuchi Huo
}
\begin{document}

\maketitle

\begin{abstract}
This paper propose a interactive 3D modeling method and corresponding system based on single or multiple uncalibrated images. The main feature of this method is that, according to the modeling habits of ordinary people, the 3D model of the target is reconstructed from coarse to fine images. On the basis of determining the approximate shape, the user adds or modify projection constraints and spatial constraints, and apply topology modification, gradually realize camera calibration, refine rough model, and finally complete the reconstruction of objects with arbitrary geometry and topology. During the interactive process, the geometric parameters and camera projection matrix are solved in real time, and the reconstruction results are displayed in a 3D window.
\end{abstract}

\section{Introduction}

As an important supplement to traditional computer-aided 3D modeling technology, image-based modeling technology has made great progress in recent years. The so-called image-based modeling (IBM) technology is to recover the geometry and surface texture and other attribute information of objects from real images through technical means, so that the created models are more visually realistic.

In IBM technology, there are two main methods: 1) Geometric reconstruction method based on discrete points. It uses the corresponding points on multiple images and adopts the principle of stereo vision to restore the three-dimensional coordinates of the points, and then establishes a polygonal mesh model through the topological connection between these points. The advantage of this method is that it can theoretically reconstruct any shape of geometry; the disadvantage is that in practical applications, for objects with complex geometry and texture, there is a problem of low reliability in automatically selecting and matching corresponding points between multiple images. However, manual matching of corresponding points has the problem of excessive interaction. 2) Modeling method based on predefined voxels. A three-dimensional geometric parameterized model of the target model is given through human-computer interaction, and then by establishing the corresponding point-line relationship between the three-dimensional model and the image and the constraint relationship of the three-dimensional geometry itself, the precise parameters of the geometric model are determined by the optimization method. The advantage of this method is that it utilizes the constraint relationship of the model itself, and the amount of interaction is smaller than the previous method; the disadvantage is that a parameterized model needs to be built in advance, which limits the geometry type of the recoverable target model.

We propose a progressive interactive image-based modeling method using a pre-defined voxel-based approach and incorporating ideas from discrete point methods. Using this modeling approach, we developed an IBM modeling system with high versatility, reliability, and ease of use.

\section{Progressive Interactive Modeling}

Image-based modeling restores the 3D model based on the information of the image. Without loss of generality, we believe that the 3D model to be restored can be represented as a grid. The mesh model M can be represented by a quadruple (K, V, D, S) \cite{r4}. Among them, V is a set of vertex positions {v1, . . . , vn}, which defines the shape of the mesh; K is the connection relationship between V and defines the topology of the mesh; D defines the discrete surface of the model attributes, such as the surface material identification number; S defines the continuous attributes indicated by the model, such as color, normal, texture coordinates, etc. The goal of image-based reconstruction is to recover such a quaternary relation M in space from the image.

Hoppe proved that the conversion between two grids $M^n$ and $M^0$ in space only needs limited steps to realize \cite{r4}. We introduce such ideas into image-based modeling techniques and propose a progressive interactive image-based modeling method. We assume that the model to be recovered from several images $I_j (j=1,...,N)$ is $M^{\infty}$, and the vertex $v_i$ must satisfy the projection constraint: $m_i^j = s P_j v_i$. Among them, $m^i_j$ is the plane point on the image $I_j$, which is a known condition; $s$ is the projection coefficient; $P_j$ is the camera projection matrix corresponding to the image $I_j$ ($m^j_i$ and $v_i$ are expressed in the form of homogeneous coordinates). What needs to be solved is $P_j$ and $v_i$. In addition to the projection constraints, the constraints that can be used in the restoration process are: epipolar constraint between cameras $f(m_i, m_i') = 0$, which defines the corresponding relationship between the plane projection points of the same space point in different images; The space constraints between vertices $g(v_{i0},...,v_{in}) = 0$, such as coplanarity, collinearity, and face-to-face parallel, face-to-face perpendicular constraints.

We represent all projection constraints, epipolar constraints and spatial constraints as a constraint relation group F. Theoretically, when there are enough constraints and certain conditions are met, the vertex set V can be recovered by solving F. In addition, if we simplify D and S to the texture and physical coordinates of the mesh surface, then when the topology K and vertex set V of the mesh are known, we can get the vertices of each face on the image plane according to the projection constraints. Through the inverse matrix of the projection matrix, we can inversely obtain the texture coordinates of the surface and the vertex. Therefore, the grid (K, V, D, S) can be obtained by the corresponding calculation of (K, F).

\begin{figure}
    \centering
    \includegraphics[width=0.45\textwidth]{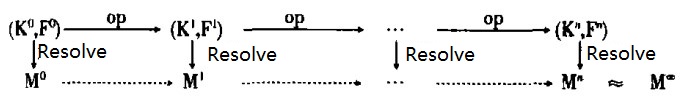}
    \caption{Progressive modeling process.}
    \label{fig1}
\end{figure}

Our modeling system can be regarded as a state machine. The user applies different operators op to the state machine to change into different states by interacting with the system, where the state can be represented by $(K^i, F^i)$, and the system can pass the pair $(K^i, F^i)$ Calculations are made to obtain mesh models $M^i$, these $M^i (i = 0,...,n)$ will approximate the final mesh $M^{\infty}$ (Figure \ref{fig1}).

The operators are divided into the following three categories according to the different imposed objects.

(1) K-op operators, that is, operators applied to topology K. It is proved in \cite{r4} that only one pair of reversible operators is needed for two grids Mn and M 0 in space: vertex splitting and edge merging. Considering that they are not conducive to intuitively controlling the grid during interaction, for this purpose, we define two pairs of equivalent reciprocal inverse operators as shown in Table \ref{tab1}.

\begin{table}[H]
    \centering
    \begin{tabular}{|c|c|}
    \hline
       $E_{split}$  &  Add a vertex to split the edge into two edges. \\ \hline
      $V_{delete}$   & The inverse operator of $E_{split}$. \\ \hline
      $V_{connect}$   & Increase a new edge to connect two vertices. \\ \hline
      $E_{delete}$   & The inverse operator of $V_{connect}$.  \\ \hline
    \end{tabular}
    \caption{Two pairs of reciprocal inverse operators.}
    \label{tab1}
\end{table}

In order to construct the mesh from an initial topology quickly, we set several initial topology construction operators ($K_{create}$) are defined to create predefined voxels.

(2) F-op operator, that is, the operator imposed on the constraint F. Two types of operators are defined: A, the projection constraint operator ($V_{proj}$), which specifies the correspondence between spatial points and pixels on the image to establish projection constraints; B, the spatial constraint operator ($V_{space}$), which specifies the spatial constraints between spatial points relation, including the corresponding inverse operator

(3) V-op operator, that is, the operator applied to the vertex set V. Although the system solves V through F, when the number of constraints in F is insufficient or when numerical iteration is used to solve V, and the initial value is not good, the system can be assisted in solving V by directly adjusting the positions of some vertices.

\begin{figure}
    \centering
    \includegraphics[width=0.45\textwidth]{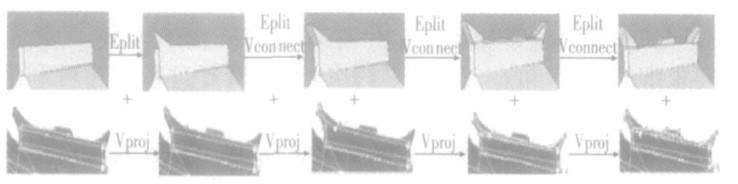}
    \caption{An example of the interactive process.}
    \label{fig2}
\end{figure}

We take the modeling process of the top of a pavilion as an example to illustrate. First, according to the top shape, use the $K_{create}$ operator to create a cuboid voxel, and then through the operator $V_{proj}$, the user drags the cube to the corresponding points under different image projections to establish projection constraints. The system uses the edge vertical information of the cube and the existing entities in the scene to calculate the camera internal parameters of each image; extracts point pairs from the corresponding relationship between multiple images to establish epipolar constraints, and calculates the camera external parameters. According to the scene's current projection constraints, spatial constraints (in hidden voxels) and updated camera parameters number, which computes the geometry of the model in the scene. Obviously, a cube cannot properly approach the top of the real scene, so the user can continue to use $E_{split}$ to split the edges of the cube and generate new vertices, as shown in Figure \ref{fig2}; The vertices need to be connected with $V_{connect}$ to complete the further approximation of the top shape of the pavilion. After each drag operation by the user, the system updates the model. Finally, the system automatically solves the texture and texture coordinates.

\section{Computational in Modeling}

\subsection{Inverse Camera Parameters}

We adopt a step-by-step refinement of the inverse algorithm of camera parameters. First, the internal parameters of the camera are recovered from a single image, and then the external parameters of the camera are recovered and the internal parameters are refined through the epipolar constraints of the corresponding points between the two images. 

When recovering the camera's internal parameters from a single image, if the user has sufficient prior knowledge of the scene in the image, as long as the known spatial points of six different planes are input, the camera's internal parameters can be reversed; otherwise, by using the space of the three-three perpendicular line clusters on the plane for the projection, the vanishing point is obtained by the method of literature \cite{r8}, and the internal parameters of the camera are reversed from the vanishing point by the method of literature \cite{r7}.

When recovering camera parameters from multiple images, the epipolar constraints between the two images are obtained through the projection relationship established by the user. Let the projection point sets of the vertex $v$ in the two images be $m$ and $m'$ respectively, and eliminate $v$, $s$, $s'$ from the equation
$\left \{
\begin{aligned}
sm=Pv\\
s'm'=P'v\\
\end{aligned}
\right.
$ (where $P$ is the camera projection matrix), and get the equation $m^{'T}Tm = 0$, that is the corresponding point epipolar constraint, where $T$ is the fundamental matrix \cite{r5,r3}. There are many ways to calculate $T$ \cite{r9}, we use the non-linear distance from the point to the limit line with the pixel distance as the final optimization method:

\begin{equation}
    \min_F \sum_i (d^2({m_i}',T m_i)+ d^2 (m_i, T {m_i}')),
\end{equation}
where $d(...)$ is the distance function. After restoring $T$, we can restore $P$ from projective transformation to affine transformation through $T$, and then directly use the aforementioned method of restoring camera internal parameters from a single image to restore $P$ from affine transformation to isometric transformation (Metric Matrix). When the number of input images is large enough, we use the self-calibration algorithm \cite{r6} to calculate the internal parameters of the camera. But in order to ensure the stability of the self-calibration algorithm, we use the internal parameters calculated from a single image as the initial value of the iteration.

\subsection{Geometry Inverse}
According to the projection constraints and space constraints, in the input $N$ images, the vertex set $V$ of the mesh is solved, which can be expressed as the following form:
\begin{equation}
    \min \sum^N_{j=0} \sum^L_{i=0} \|P_j v_i - m_i^j\|^2 \quad s.t. g(v_0,...,v_L)=0,
    \label{eq1}
\end{equation}

where $P_j$ is the camera matrix of the recovered image $I_j$, $m_j^i$ is the projection of the vertex $v_i$ of $I_j$, and $g$ is the space constraint, which is the most basic point projection expression. If the point-based expression is solved, the amount of calculation is very large, and the corresponding relationship needs to be input point by point. Therefore, a parameterized voxel representation \cite{r2} is introduced into the system, which expresses the vertex position as the local coordinate of the voxel and the product of the voxel's rotation and translation transformation. After the initial position of the voxel is calculated, according to the K-op operator used by the user for topology modification, the newly generated vertex is turned to the method based on local coordinates to participate in the solution of formula \ref{eq1}.

\subsection{Resolving Texture and Texture Coordinate}

After completing the inverse calculation of the camera parameters and the inverse calculation of the model geometry, the texture attributes of each facet of the model can be obtained. The texture inverse operation is actually a mapping transformation $T$ from the source space to the target space. $T$ can be obtained by solving the inverse of a projective transformation. Since each surface has different projected textures on different images, and there are influences such as occlusion and shadow, the colors of the projected points of the same spatial point on different images may be different. We comprehensively consider the visibility of the patch in each image, the angle between the normal direction of the patch and the line of sight of each camera, and the size of the projected area of the patch in the image. Restore high-quality viewpoint-based textures for units \cite{r1,r2}.

\section{Reconstruction Result}
\begin{figure}
    \centering
    \includegraphics[width=0.48\textwidth]{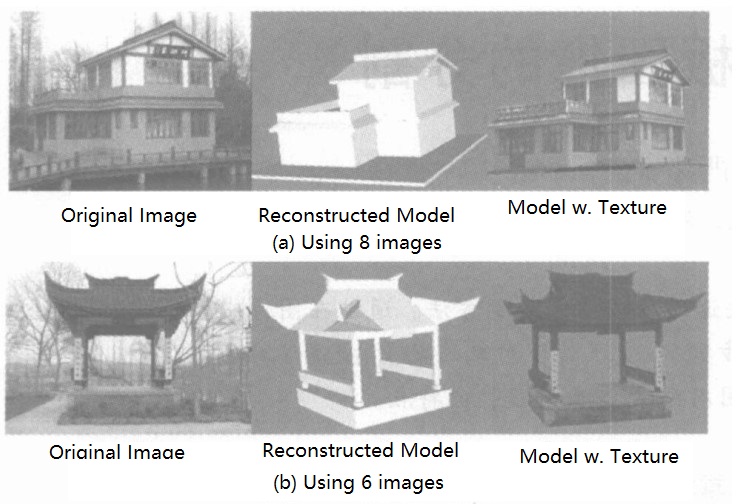}
    \caption{Model reconstruction results.}
    \label{fig3}
\end{figure}

The system is divided into four modules: interface module, space modeling module, plane image processing module and optimization solution module. Figure \ref{fig3} presents two examples of modeling by the system from multiple images.

\section{Conclusion}

We implement a progressive interactive image-based modeling approach with a corresponding system. Compared with the existing IBM method and system, the system better resolves the contradiction between interaction volume, reliability and flexibility of modeling, and creatively introduces the progressive IBM modeling method, which simplifies the modeling process. process and make it conform to human cognitive habits. The system has the following characteristics: progressive interactive solution mode; WYSIWYG interface environment; accurate camera calibration algorithm; texture editing function and accurate texture acquisition. Further work includes: on the basis of reconstructing geometry and restoring texture attributes, how to further restore the surface lighting attributes, and restore the light distribution and attributes in the scene.

There are some extended articles about applying physical lighting computation in various applications:

\begin{enumerate}
    \item Deep Learning-Based Monte Carlo Noise Reduction By training a neural network denoiser through offline learning, it can filter noisy Monte Carlo rendering results into high-quality smooth output, greatly improving physics-based Availability of rendering techniques \cite{huo2021survey}, common research includes predicting a filtering kernel based on g-buffer \cite{bako2017kernel}, using GAN to generate more realistic filtering results \cite{xu2019adversarial}, and analyzing path space features Perform manifold contrastive learning to enhance the rendering effect of reflections \cite{cho2021weakly}, use weight sharing to quickly predict the rendering kernel to speed up reconstruction \cite{fan2021real}, filter and reconstruct high-dimensional incident radiation fields for unbiased reconstruction rendering guide \cite{huo2020adaptive}, etc.
    \item The multi-light rendering framework is an important rendering framework outside the path tracing algorithm. Its basic idea is to simplify the simulation of the complete light path illumination transmission after multiple refraction and reflection to calculate the direct illumination from many virtual light sources, and provide a unified Mathematical framework to speed up this operation \cite{dachsbacher2014scalable}, including how to efficiently process virtual point lights and geometric data in external memory \cite{wang2013gpu}, how to efficiently integrate virtual point lights using sparse matrices and compressed sensing \cite{huo2015matrix}, and how to handle virtual line light data in translucent media \cite{huo2016adaptive}, use spherical Gaussian virtual point lights to approximate indirect reflections on glossy surfaces \cite{huo2020spherical}, and more.
    \item Automatic optimization of rendering pipelines Apply high-quality rendering technology to real-time rendering applications by optimizing rendering pipelines. The research contents include automatic optimization based on quality and speed \cite{wang2014automatic}, automatic optimization for energy saving \cite{ wang2016real,zhang2021powernet}, LOD optimization for terrain data \cite{li2021multi}, automatic optimization and fitting of pipeline rendering signals \cite{li2020automatic}, anti-aliasing \cite{zhong2022morphological}, etc.
    \item Using physically-based process to guide the generation of data for single image reflection removal \cite{kim2020single}; propagating local image features in a hypergraph for image retreival \cite{an2021hypergraph}; managing 3D assets in a block chain-based distributed system \cite{park2021meshchain}.
\end{enumerate}

% many light method, VPL

\bibliographystyle{ieee}
\bibliography{srbib}

\end{document}